\documentclass[11pt,letterpaper]{article}
\usepackage{emnlp2016}
\usepackage{times}
\usepackage{latexsym}

\usepackage{amsmath}
\usepackage{amssymb}
\usepackage{amsopn}
\usepackage{latexsym}
\usepackage{subcaption}
\usepackage{CJKutf8}
\usepackage{graphicx}
\usepackage{multirow}
\usepackage{color}
\usepackage[usenames,dvipsnames]{xcolor}
\usepackage{tabularx}
\usepackage{algorithm}  
\usepackage{algorithmic}  
\usepackage[skip=2pt,belowskip=-12pt,aboveskip=2pt,font=small]{caption}

\definecolor{zb_dred}{RGB}{102, 0, 0}
\definecolor{zb_red}{RGB}{255, 0, 0}
\definecolor{zb_lred}{RGB}{255, 102, 102} 

\usepackage{enumitem}
\setenumerate[1]{itemsep=0.30pt,partopsep=0pt,parsep=\parskip,topsep=5pt}
\setitemize[1]{itemsep=0.30pt,partopsep=00pt,parsep=\parskip,topsep=5pt}
\setdescription{itemsep=0.30pt,partopsep=0pt,parsep=\parskip,topsep=5pt}

\emnlpfinalcopy

\def\emnlppaperid{***}

\title{Variational Neural Discourse Relation Recognizer}

\author{Biao Zhang$^{1}$, Deyi Xiong$^{2}$\thanks{~~Corresponding author}, Jinsong Su$^{1}$, Qun Liu$^{3,4}$, Rongrong Ji$^{1}$, Hong Duan$^{1}$, Min Zhang$^{2}$\\
	Xiamen University, Xiamen, China 361005$^{1}$ \\
	Provincial Key Laboratory for Computer Information Processing Technology \\
	Soochow University, Suzhou, China 215006$^{2}$ \\
	ADAPT Centre, School of Computing, Dublin City University$^{3}$ \\
	Key Laboratory of Intelligent Information Processing, \\
	Institute of Computing Technology, Chinese Academy of Sciences$^{4}$ \\
	{\tt zb@stu.xmu.edu.cn, \{jssu, rrji, hduan\}@xmu.edu.cn} \\
	{\tt qun.liu@dcu.ie, \{dyxiong, minzhang\}@suda.edu.cn} \\
}

\date{}

\begin{document}

\maketitle

\begin{abstract}

Implicit discourse relation recognition is a crucial component for automatic discourse-level analysis and nature language understanding. Previous studies exploit discriminative models that are built on either powerful manual features or deep discourse representations. In this paper, instead, we explore generative models and propose a variational neural discourse relation recognizer. We refer to this model as {\it VarNDRR}. VarNDRR establishes a directed probabilistic model with a latent continuous variable that generates both a discourse and the relation between the two arguments of the discourse. In order to perform efficient inference and learning, we introduce neural discourse relation models to approximate the prior and posterior distributions of the latent variable, and employ these approximated distributions to optimize a reparameterized variational lower bound. This allows VarNDRR to be trained with standard stochastic gradient methods. Experiments on the benchmark data set show that VarNDRR can achieve comparable results against state-of-the-art baselines without using any manual features.

\end{abstract}

\section{Introduction}

Discourse relation characterizes the internal structure and logical relation of a coherent text. Automatically identifying these relations not only plays an important role in discourse comprehension and generation, but also obtains wide applications in many other relevant natural language processing tasks, such as text summarization \cite{yoshida-EtAl:2014:EMNLP2014}, conversation \cite{conf/coling/HigashinakaIMMKSHMM14}, question answering \cite{Verberne:2007:EDA:1277741.1277883} and information extraction \cite{cimiano2005ontology}. Generally, discourse relations can be divided into two categories: explicit and implicit, which can be illustrated in the following example:
\begin{quote}
{\it The company was disappointed by the ruling.}
{\it \underline{because} The obligation is totally unwarranted.}
(adapted from wsj\_0294)
\end{quote}
With the discourse connective {\it because}, these two sentences display an explicit discourse relation \textsc{Contingency} which can be inferred easily. Once this discourse connective is removed, however, the discourse relation becomes implicit and difficult to be recognized. This is because almost no surface information in these two sentences can signal this relation. For successful recognition of this relation, in the contrary, we need to understand the deep semantic correlation between {\it disappointed} and {\it obligation} in the two sentences above. Although explicit discourse relation recognition (DRR) has made great progress \cite{miltsakaki2005experiments,pitler2008easily}, implicit DRR still remains a serious challenge due to the difficulty in semantic analysis.

Conventional approaches to implicit DRR often treat the relation recognition as a classification problem, where discourse arguments and relations are regarded as the inputs and outputs respectively. Generally, these methods first generate a representation for a discourse, denoted as $\mathbf{x}$\footnote{Unless otherwise specified, all variables in the paper, e.g., $\mathbf{x},\mathbf{y},\mathbf{z}$ are multivariate. But for notational convenience, we treat them as univariate variables in most cases. Additionally, we use bold symbols to denote variables, and plain symbols to denote values.} (e.g., manual features in SVM-based recognition \cite{pitler-louis-nenkova:2009:ACLIJCNLP,lin2009recognizing} or sentence embeddings in neural networks-based recognition \cite{TACL536,biaozhang:2015:emnlp:drr}), and then directly model the conditional probability of the corresponding discourse relation $\mathbf{y}$ given $\mathbf{x}$, i.e. $p(\mathbf{y}|\mathbf{x})$. In spite of their success, these discriminative approaches rely heavily on the goodness of discourse representation $\mathbf{x}$. Sophisticated and good representations of a discourse, however, may make models suffer from overfitting as we have no large-scale balanced data.	

Instead, we assume that there is a latent continuous variable $\mathbf{z}$ from an underlying semantic space. It is this latent variable that generates both discourse arguments and the corresponding relation, i.e. $p(\mathbf{x},\mathbf{y}|\mathbf{z})$. The latent variable enables us to jointly model discourse arguments and their relations, rather than conditionally model $\mathbf{y}$ on $\mathbf{x}$. However, the incorporation of the latent variable makes the modeling difficult due to the intractable computation with respect to the posterior distribution.

\begin{figure}[t]
\centering
\includegraphics[scale=0.80]{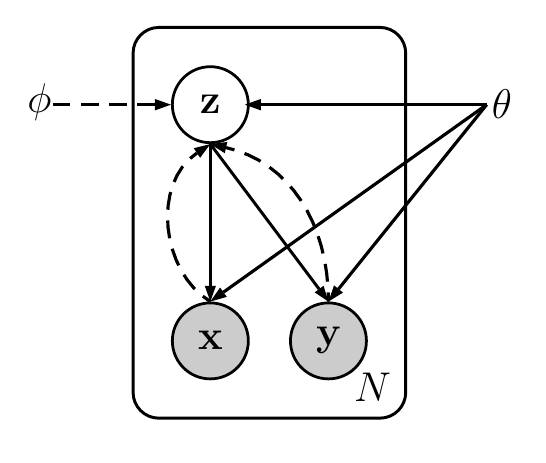}
\caption{\label{graph_model} Graphical illustration for VarNDRR. Solid lines denote the generative model $p_\theta(\mathbf{x}|\mathbf{z})p_\theta(\mathbf{y}|\mathbf{z})$, dashed lines denote the variational approximation $q_\phi(\mathbf{z}|\mathbf{x}, \mathbf{y})$ to the posterior $p(\mathbf{z}|\mathbf{x}, \mathbf{y})$ and $q_\phi^{\prime}(\mathbf{z}|\mathbf{x})$ to the prior $p(\mathbf{z})$ for inference. The variational parameters $\phi$ are learned jointly with the generative model parameters $\theta$.}
\end{figure}

Inspired by Kingma and Welling~\shortcite{kingma2014autoencoding} as well as Rezende et al.~\shortcite{DBLP:conf/icml/RezendeMW14} who introduce a variational neural inference model to the intractable posterior via optimizing a reparameterized variational lower bound, we propose a variational neural discourse relation recognizer (VarNDRR) with a latent continuous variable for implicit DRR in this paper. The key idea behind VarNDRR is that although the posterior distribution is intractable, we can approximate it via a deep neural network. Figure \ref{graph_model} illustrates the graph structure of VarNDRR. Specifically, there are two essential components: 
\begin{itemize}
\item 
{\it neural discourse recognizer} As a discourse $\mathbf{x}$ and its corresponding relation $\mathbf{y}$ are independent with each other given the latent variable $\mathbf{z}$ (as shown by the solid lines), we can formulate the generation of $\mathbf{x}$ and $\mathbf{y}$ from $\mathbf{z}$ in the equation $p_\theta(\mathbf{x},\mathbf{y}|\mathbf{z})=p_\theta(\mathbf{x}|\mathbf{z})p_\theta(\mathbf{y}|\mathbf{z})$. These two conditional probabilities on the right hand side are modeled via deep neural networks (see section \ref{recognizer}).
\item
{\it neural latent approximator} VarNDRR assumes that the latent variable can be inferred from discourse arguments $\mathbf{x}$ and relations $\mathbf{y}$ (as shown by the dash lines). In order to infer the latent variable, we employ a deep neural network to approximate the posterior $q_\phi(\mathbf{z}|\mathbf{x}, \mathbf{y})$ as well as the prior $q_\phi^{\prime}(\mathbf{z}|\mathbf{x})$ (see section \ref{approximator}), which makes the inference procedure efficient. 
We further employ a reparameterization technique to sample $z$ from $q_\phi(\mathbf{z}|\mathbf{x}, \mathbf{y})$ that not only bridges the gap between the recognizer and the approximator but also allows us to use the standard stochastic gradient ascent techniques for optimization (see section \ref{learning}).
\end{itemize}
The main contributions of our work lie in two aspects. 1) We exploit a generative graphic model for implicit DRR. To the best of our knowledge, this has never been investigated before. 2) We develop a neural recognizer and two neural approximators specifically for implicit DRR, which enables both the recognition and inference to be efficient. 

We conduct a series of experiments for English implicit DRR on the PDTB-style corpus to evaluate the effectiveness of our proposed VarNDRR model. Experiment results show that our variational model achieves comparable results against several strong baselines in term of F1 score. Extensive analysis on the variational lower bound further reveals that our model can indeed fit the data set with respect to discourse arguments and relations.

\section{Background: Variational Autoencoder}

The variational autoencoder (VAE)~\cite{kingma2014autoencoding,DBLP:conf/icml/RezendeMW14}, which forms the basis of our model, 
is a generative model that can be regarded as a regularized version of the standard autoencoder. With a latent random variable $\mathbf{z}$, VAE significantly changes the autoencoder architecture to be able to capture the variations in the observed variable $\mathbf{x}$. 
The joint distribution of $(\mathbf{x}, \mathbf{z})$ is formulated as follows:
\begin{equation}\label{vae_joint}
p_\theta(\mathbf{x}, \mathbf{z}) = p_\theta(\mathbf{x}|\mathbf{z})p_\theta(\mathbf{z})
\end{equation}
where $p_\theta(\mathbf{z})$ is the prior over the latent variable, usually equipped with a simple Gaussian distribution. $p_\theta(\mathbf{x}|\mathbf{z})$ is the conditional distribution that models the probability of $\mathbf{x}$ given the latent variable $\mathbf{z}$. Typically, VAE parameterizes $p_\theta(\mathbf{x}|\mathbf{z})$ with a highly non-linear but flexible function approximator such as a neural network.

The objective of VAE is to maximize a variational lower bound as follows:
\begin{equation}\label{vae_vlb}
\begin{split}
\mathcal{L}_{VAE}(\theta,\phi;\mathbf{x}) = -&\text{KL}(q_\phi(\mathbf{z}|\mathbf{x})||p_\theta(\mathbf{z})) \\
+ \mathbb{E}_{q_\phi(\mathbf{z}|\mathbf{x})}&[\log p_\theta(\mathbf{x}|\mathbf{z})] \leq \log p_\theta(\mathbf{x})
\end{split}
\end{equation}
where $\text{KL}(Q||P)$ is Kullback-Leibler divergence between two distributions $Q$ and $P$. $q_\phi(\mathbf{z}|\mathbf{x})$ is an approximation of the posterior $p(\mathbf{z}|\mathbf{x})$ and usually follows a diagonal Gaussian $\mathcal{N}(\mathbf{\mu}, \text{diag}(\mathbf{\sigma^2}))$ whose mean $\mathbf{\mu}$ and variance $\mathbf{\sigma^2}$ are parameterized by again, neural networks, conditioned on $\mathbf{x}$.


To optimize Eq. (\ref{vae_vlb}) stochastically with respect to both $\theta$ and $\phi$, VAE introduces a reparameterization trick that parameterizes the latent variable $\mathbf{z}$ with the Gaussian parameters $\mathbf{\mu}$ and $\mathbf{\sigma}$ in $q_\phi(\mathbf{z}|\mathbf{x})$:
\begin{equation}\label{vae_re_param}
\tilde{{z}} = \mathbf{\mu} + \mathbf{\sigma}\odot \mathbf{\epsilon}
\end{equation}
where $\mathbf{\epsilon}$ is a standard Gaussian variable, and $\odot$ denotes an element-wise product. Intuitively, VAE learns the representation of the latent variable not as single points, but as soft ellipsoidal regions in latent space, forcing the representation to fill the space rather than memorizing the training data as isolated representations. With this trick, the VAE model can be trained through standard backpropagation technique with stochastic gradient ascent.

\section{The VarNDRR Model}

This section introduces our proposed VarNDRR model. Formally, in VarNDRR, there are two observed variables, $\mathbf{x}$ for a discourse and $\mathbf{y}$ for the corresponding relation, and one latent variable $\mathbf{z}$. As illustrated in Figure \ref{graph_model}, the joint distribution of the three variables is formulated as follows:
\begin{equation} \label{our_joint}
p_\theta(\mathbf{x}, \mathbf{y}, \mathbf{z}) = p_\theta(\mathbf{x}, \mathbf{y}|\mathbf{z})p(\mathbf{z})
\end{equation}
We begin with this distribution to elaborate the major components of VarNDRR.

\subsection{Neural Discourse Recognizer} \label{recognizer}

\begin{figure}[t]
\centering
\includegraphics[scale=0.80]{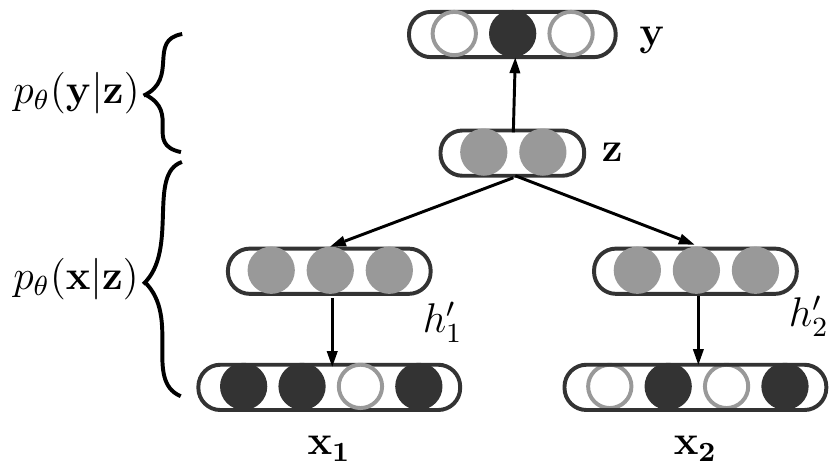}
\caption{\label{our_recognizer} Neural networks for conditional probabilities $p_\theta(\mathbf{x}|\mathbf{z})$ and $p_\theta(\mathbf{y}|\mathbf{z})$. The gray color denotes real-valued representations while the white and black color 0-1 representations.}
\end{figure}

The conditional distribution $p(\mathbf{x}, \mathbf{y}|\mathbf{z})$ in Eq. (\ref{our_joint}) shows that both discourse arguments and the corresponding relation are generated from the latent variable. 
As shown in Figure \ref{graph_model}, $\mathbf{x}$ is d-separated from $\mathbf{y}$ by $\mathbf{z}$. Therefore the discourse $\mathbf{x}$ and the corresponding relation $\mathbf{y}$ is independent given the latent variable $\mathbf{z}$. The joint probability can be therefore formulated as follows:
\begin{equation} \label{our_joint_new}
p_\theta(\mathbf{x}, \mathbf{y}, \mathbf{z}) = p_\theta(\mathbf{x}|\mathbf{z})p_\theta(\mathbf{y}|\mathbf{z})p(\mathbf{z})
\end{equation}
We use a neural model $q_\phi^{\prime}(\mathbf{z}|\mathbf{x})$ to approximate the prior $p(\mathbf{z})$ conditioned on the discourse $\mathbf{x}$ (see the following section).
With respect to the other two conditional distributions, we parameterize them via neural networks as shown in Figure \ref{our_recognizer}.

Before we describe these neural networks, it is necessary to briefly introduce how discourse relations are annotated in our training data. The PDTB corpus, used as our training data, annotates implicit discourse relations between two neighboring arguments, namely {\it Arg1} and {\it Arg2}. In VarNDRR, we represent the two arguments with bag-of-word representations, and denote them as $\mathbf{x_1}$ and $\mathbf{x_2}$. 

To model $p_\theta(\mathbf{x}|\mathbf{z})$ (the bottom part in Figure \ref{our_recognizer}), we project the representation of the latent variable $z\in \mathbb{R}^{d_z}$ onto a hidden layer:
\begin{equation}
\begin{split}
h_{1}^{\prime} = f(W_{h_{1}^{\prime}} z + b_{h_{1}^{\prime}})\\
h_{2}^{\prime} = f(W_{h_{2}^{\prime}} z + b_{h_{1}^{\prime}})
\end{split}
\end{equation}
where $h_{1}^{\prime} \in \mathbb{R}^{d_{h_{1}^{\prime}}}, h_{2}^{\prime} \in \mathbb{R}^{d_{h_{2}^{\prime}}}$, $W_{*}$ is the transformation matrix, $b_{*}$ is the bias term, $d_u$ denotes the dimensionality of vector representations of $u$ and $f(\cdot)$ is an element-wise activation function, such as $tanh(\cdot)$, which is used throughout our model. 

Upon this hidden layer, we further stack a Sigmoid layer to predict the probabilities of corresponding discourse arguments:
\begin{equation}
\begin{split}
x_{1}^{\prime} = \text{Sigmoid}(W_{x_{1}^{\prime}} h_{1}^{\prime} + b_{x_{1}^{\prime}}) \\
x_{2}^{\prime} = \text{Sigmoid}(W_{x_{2}^{\prime}} h_{2}^{\prime} + b_{x_{2}^{\prime}})
\end{split}
\end{equation}
here, $x_{1}^{\prime} \in \mathbb{R}^{d_{x_{1}}}$ and $x_{2}^{\prime} \in \mathbb{R}^{d_{x_{2}}}$ are the real-valued representations of the reconstructed $x_1$ and $x_2$ respectively.\footnote{Notice that the equality of $d_{x_{1}} = d_{x_2}, d_{h_{1}^{\prime}}=d_{h_{2}^{\prime}}$ is not necessary though we assume so in our experiments.} We assume that $p_\theta(\mathbf{x}|\mathbf{z})$ is a multivariate Bernoulli distribution because of the bag-of-word representation. Therefore the logarithm of $p(x|z)$ is calculated as the sum of probabilities of words in discourse arguments as follows:
\begin{equation}\label{log_sum_pxz}
\begin{split}
&\log ~p(x|z) \\
&=\sum_{i} x_{1,i}\log x_{1,i}^{\prime} + (1-x_{1,i})\log (1- x_{1,i}^{\prime}) \\
&+\sum_{j} x_{2,j}\log x_{2,j}^{\prime} + (1-x_{2,j})\log (1- x_{2,j}^{\prime}) 
\end{split}
\end{equation}
where $u_{i,j}$ is the $j$th element in $u_{i}$.

In order to estimate $p_\theta(\mathbf{y}|\mathbf{z})$ (the top part in Figure \ref{our_recognizer}), we stack a softmax layer over the multilayer-perceptron-transformed representation of the latent variable $z$:
\begin{equation}
y^{\prime} = \text{SoftMax}(W_{y^{\prime}} \text{MLP}(z) + b_{y^{\prime}})
\end{equation}
$y^{\prime} \in \mathbb{R}^{d_{y}}$, and $d_{y}$ denotes the number of discourse relations. MLP projects the representation of latent variable $\mathbf{z}$ into a $d_m$-dimensional space through four internal layers, each of which has dimension $d_m$. Suppose that the true relation is $y \in \mathbb{R}^{d_y}$, the logarithm of $p(y|z)$ is defined as:
\begin{equation}\label{log_sum_pyz}
\log~p(y|z) = \sum_{i=1}^{d_y} y_i \log y_{i}^{\prime}
\end{equation}

In order to precisely estimate these conditional probabilities, our model will force the representation $z$ of the latent variable to encode semantic information for both the reconstructed discourse $x^\prime$ (Eq. (\ref{log_sum_pxz})) and predicted discourse relation $y^\prime$ (Eq. (\ref{log_sum_pyz})), which is exactly what we want.


\subsection{Neural Latent Approximator} \label{approximator}

\begin{figure}[t]
\centering
\includegraphics[scale=0.80]{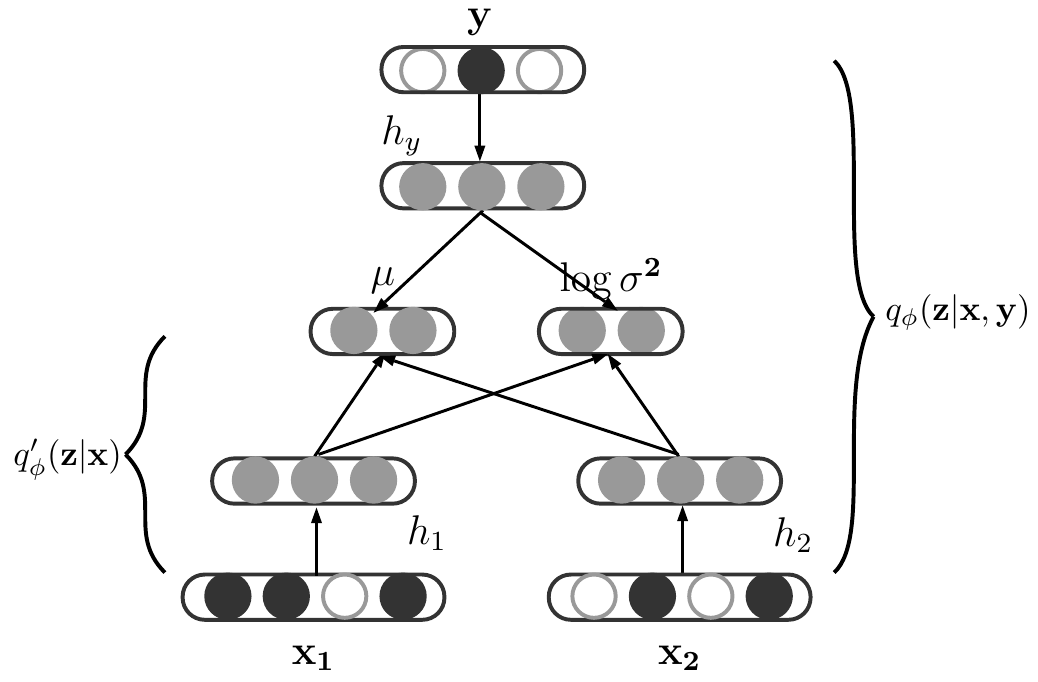}
\caption{\label{our_posterior} Neural networks for Gaussian parameters $\mathbf{\mu}$ and $\mathbf{\log \sigma}$ in the approximated posterior $q_\phi(\mathbf{z}|\mathbf{x}, \mathbf{y})$ and prior $q_\phi^{\prime}(\mathbf{z}|\mathbf{x})$.}
\end{figure}

For the joint distribution in Eq. (\ref{our_joint_new}), we can define a variational lower bound that is similar to Eq. (\ref{vae_vlb}). The difference lies in two aspects: the approximate prior $q_\phi^{\prime}(\mathbf{z}|\mathbf{x})$ and posterior $q_\phi(\mathbf{z}|\mathbf{x}, \mathbf{y})$. We model both distributions as a multivariate Gaussian distribution with a diagonal covariance structure:
\begin{equation*}
\mathcal{N}(\mathbf{z};\mathbf{\mu}, \mathbf{\sigma}^2\mathbf{I})
\end{equation*}
The mean $\mathbf{\mu}$ and s.d. $\mathbf{\sigma}$ of the approximate distribution are the outputs of the neural network as shown in Figure \ref{our_posterior}, where the prior and posterior have different conditions and independent parameters.


{\it Approximate Posterior} $q_\phi(\mathbf{z}|\mathbf{x}, \mathbf{y})$ is modeled to condition on both observed variables: the discourse arguments $\mathbf{x}$ and relations $\mathbf{y}$. Similar to the calculation of $p_\theta(\mathbf{x}|\mathbf{z})$, we first transform the input $\mathbf{x}$ and $\mathbf{y}$ into a hidden representation:
\begin{equation}\label{posterior_hidden}
\begin{split}
&h_1 = f(W_{h_1} x_1 + b_{h_1}) \\
&h_2 = f(W_{h_2} x_2 + b_{h_2}) \\
&h_y = f(W_{h_y} y + b_{h_y})
\end{split}
\end{equation}
where $h_1 \in \mathbb{R}^{d_{h_1}}, h_2 \in \mathbb{R}^{d_{h_2}}$ and $h_y \in \mathbb{R}^{d_{h_y}}$.\footnote{Notice that $d_{h_1}/d_{h_2}$ are not necessarily equal to $d_{h_{1}^{\prime}}/d_{h_{2}^{\prime}}$.}

We then obtain the Gaussian parameters of the posterior $\mathbf{\mu}$ and $\mathbf{\log \sigma}^2$ through linear regression:
\begin{equation}\label{posterior_mu_sigma}
\begin{split}
\mu = W_{\mu_1} h_1 + W_{\mu_2} h_2 + W_{\mu_y} h_y + b_{\mu} \\
\log \sigma^2 = W_{\sigma_1} h_1 + W_{\sigma_2} h_2 + W_{\sigma_y} h_y + b_{\sigma}
\end{split}
\end{equation}
where $\mu,\sigma \in \mathbb{R}^{d_z}$. In this way, this posterior approximator can be efficiently computed.

{\it Approximate Prior} $q_\phi^{\prime}(\mathbf{z}|\mathbf{x})$ is modeled to condition on discourse arguments $\mathbf{x}$ alone. This is based on the observation that discriminative models are able to obtain promising results using only $\mathbf{x}$. Therefore, assuming the discourse arguments encode the prior information for discourse relation recognition is meaningful.

The neural model for prior $q_\phi^{\prime}(\mathbf{z}|\mathbf{x})$ is the same as that (i.e. Eq (\ref{posterior_hidden}) and (\ref{posterior_mu_sigma})) for posterior $q_\phi(\mathbf{z}|\mathbf{x}, \mathbf{y})$ (see Figure \ref{our_posterior}), except for the absence of discourse relation $\mathbf{y}$. For clarity , we use $\mu^{\prime}$ and $\sigma^{\prime}$ to denote the mean and s.d. of the approximate prior.

With the parameters of Gaussian distribution, we can access the representation $z$ using different sampling strategies. However, traditional sampling approaches often breaks off the connection between recognizer and approximator, making the optimization difficult. Instead, we employ the reparameterization trick~\cite{kingma2014autoencoding,DBLP:conf/icml/RezendeMW14} as in Eq. (\ref{vae_re_param}). During training, we sample the latent variable using $\tilde{z}=\mu + \sigma \odot \epsilon$; during testing, however, we employ the expectation of $\mathbf{z}$ in the approximate prior distribution, i.e. set $\tilde{z}=\mu^{\prime}$ to avoid uncertainty. 

%

\subsection{Parameter Learning} \label{learning}

We employ the Monte Carlo method to estimate the expectation over the approximate posterior, that is $\mathbb{E}_{q_\phi(\mathbf{z}|\mathbf{x}, \mathbf{y})}[\log p_\theta(\mathbf{x},\mathbf{y}|\mathbf{z})]$. Given a training instance $(x^{(t)}, y^{(t)})$, the joint training objective is defined: 
\begin{align} \label{final_object}
\mathcal{L}(\theta, \phi) \simeq -\text{KL}&(q_\phi(\mathbf{z}|x^{(t)}, y^{(t)})||q_\phi^{\prime}(\mathbf{z}|x^{(t)}))   \notag \\
\qquad + \frac{1}{L}&\sum_{l=1}^L \log p_\theta(x^{(t)},y^{(t)}|\tilde{z}^{(t,l)}) \\
\text{where}~\tilde{z}^{(t,l)} = \mu^{(t)}& + \sigma^{(t)} \odot \epsilon^{(l)}~\text{and}~\epsilon^{(l)}\sim \mathcal{N}(0,\mathbf{I}) \notag
\end{align}
$L$ is the number of samples. The first term is the $\text{KL}$ divergence of two Gaussian distributions which can be computed and differentiated without estimation. Maximizing this objective will minimize the difference between the approximate posterior and prior, thus making the setting $\tilde{z}=\mu^{\prime}$ during testing reasonable. The second term is the approximate expectation of $\mathbb{E}_{q_\phi(\mathbf{z}|\mathbf{x}, \mathbf{y})}[\log p_\theta(\mathbf{x},\mathbf{y}|\mathbf{z})]$, which is also differentiable. 

As the objective function in Eq. (\ref{final_object}) is differentiable, we can optimize both the model parameters $\theta$ and variational parameters $\phi$ jointly using standard gradient ascent techniques. The training procedure for VarNDRR is summarized in Algorithm \ref{algorithm}. 

\begin{algorithm}[t]
\small
\caption{Parameter Learning Algorithm of VarNDRR.}
\label{algorithm}
\begin{algorithmic}
\STATE {Inputs: $A$, the maximum number of iterations;}
\STATE {~~~~~~~~~~~ $M$, the number of instances in one batch;}
\STATE {~~~~~~~~~~~ $L$, the number of samples;}
\STATE {$\theta,\phi$ $\leftarrow$ Initialize parameters}
\REPEAT
\STATE {$\mathcal{D}$ $\leftarrow$ getRandomMiniBatch(M)}
\STATE {$\epsilon$ $\leftarrow$ getRandomNoiseFromStandardGaussian()}
\STATE {$g$ $\leftarrow$ $\nabla_{\theta,\phi}\mathcal{L}(\theta,\phi;\mathcal{D},\epsilon)$}
\STATE {$\theta,\phi$ $\leftarrow$ parameterUpdater($\theta,\phi;g$)}
\UNTIL {convergence of parameters $(\theta,\phi)$ or reach the maximum iteration $A$}
\end{algorithmic}
\end{algorithm}

\begin{table}[t]
\centering
\small
\begin{tabular}{c|c|c|c}
\hline
\multirow{2}{*}{\bf Relation} & \multicolumn{3}{c}{\bf \#Instance Number} \\
\cline{2-4}
 & {\bf Train} & {\bf Dev} & {\bf Test} \\
\hline
\hline
\textsc{Com} & 1942 & 197 & 152 \\
\textsc{Con} & 3342 & 295 & 279 \\
\textsc{Exp} & 7004 & 671 & 574 \\
\textsc{Tem} & 760 & 64 & 85 \\
\hline		
\end{tabular}
\caption{\label{pdtb_data} Statistics of implicit discourse relations for the training (Train), development (Dev) and test (Test) sets in PDTB.}
\end{table}

\begin{table*}[t]
\centering
\small
\captionsetup[subtable]{skip=12pt,belowskip=3pt,aboveskip=6pt}
\begin{subtable}{.43\textwidth}
\centering
\begin{tabular}{c|c|c|c|c}
\hline
{\bf Model} &  {\bf Acc} & {\bf P} & {\bf R} & {\bf F1} \\
\hline
\hline
\bf R $\&$ X~\shortcite{rutherford-xue:2015:NAACL-HLT} & - & - & - & {41.00} \\
\bf J $\&$ E~\shortcite{TACL536} & 70.27 & - & - & 35.93 \\
\hline
\hline
{\bf SVM} & 63.10 & 22.79 & 64.47 & {33.68} \\ 

{\bf SCNN} & 60.42 & 22.00 & 67.76 & 33.22 \\ 
\hline
\hline
{\bf VarNDRR} & 63.30 & 24.00 & 71.05 & 35.88 \\
\hline
\end{tabular}

\caption{\textsc{Com} vs Other}
\end{subtable}
\quad\quad~~~
\begin{subtable}{.43\textwidth}
\centering
\begin{tabular}{c|c|c|c|c}
\hline
{\bf Model} & {\bf Acc} & {\bf P} & {\bf R} & {\bf F1} \\
\hline
\hline
\bf (R $\&$ X~\shortcite{rutherford-xue:2015:NAACL-HLT}) & - & - & - & {53.80} \\
\bf (J $\&$ E~\shortcite{TACL536}) & 76.95 & - & - & 52.78 \\
\hline
\hline
{\bf SVM} & 62.62 & 39.14 & 72.40 & 50.82 \\

{\bf SCNN}	& 63.00 & 39.80 & 75.29 & {52.04} \\ 
\hline
\hline
{\bf VarNDRR} & 53.82 & 35.39 & 88.53 & 50.56 \\
\hline
\end{tabular}

\caption{\textsc{Con} vs Other}
\end{subtable}
\quad
\begin{subtable}{.43\textwidth}
\centering

\begin{tabular}{c|c|c|c|c}
\hline
{\bf Model} & {\bf Acc} &  {\bf P} & {\bf R} & {\bf F1} \\
\hline
\hline
\bf (R $\&$ X~\shortcite{rutherford-xue:2015:NAACL-HLT}) & - & - & - & 69.40 \\
\bf (J $\&$ E~\shortcite{TACL536}) & 69.80 & - & - & {80.02} \\
\hline
\hline
{\bf SVM} & 60.71 & 65.89 & 58.89 & 62.19 \\

{\bf SCNN}	& 63.00 & 56.29 & 91.11 & 69.59 \\ 
\hline
\hline
{\bf VarNDRR} & 57.36 & 56.46 & 97.39 & 71.48 \\
\hline
\end{tabular}

\caption{\textsc{Exp} vs Other}
\end{subtable}
\quad\quad~~~
\begin{subtable}{.43\textwidth}
\centering

\begin{tabular}{c|c|c|c|c}
\hline
{\bf Model} & {\bf Acc} & {\bf P} & {\bf R} & {\bf F1} \\
\hline
\hline
\bf (R $\&$ X~\shortcite{rutherford-xue:2015:NAACL-HLT}) & - & - & - & {33.30} \\
\bf (J $\&$ E~\shortcite{TACL536}) & 87.11 & - & - & 27.63 \\
\hline
\hline
{\bf SVM} & 66.25 & 15.10 & 68.24 & 24.73 \\

{\bf SCNN}	& 76.95 & 20.22 & 62.35 & {30.54} \\ 
\hline
\hline
{\bf VarNDRR} & 62.14 & 17.40 & 97.65 & 29.54 \\
\hline
\end{tabular}

\caption{\textsc{Tem} vs Other}
\end{subtable}

\caption{\label{class_result} Classification results of different models on the  implicit DRR task. {\bf Acc}=Accuracy, {\bf P}=Precision, {\bf R}=Recall, and {\bf F1}=F1 score.}
\end{table*}

\begin{figure*}[t]
  \centering
  \captionsetup[subfigure]{skip=12pt,belowskip=3pt,aboveskip=6pt}
  \begin{subtable}{.33\textwidth}
  	\centering
  	\begin{tabular}{c}
  		\includegraphics[width=\textwidth]{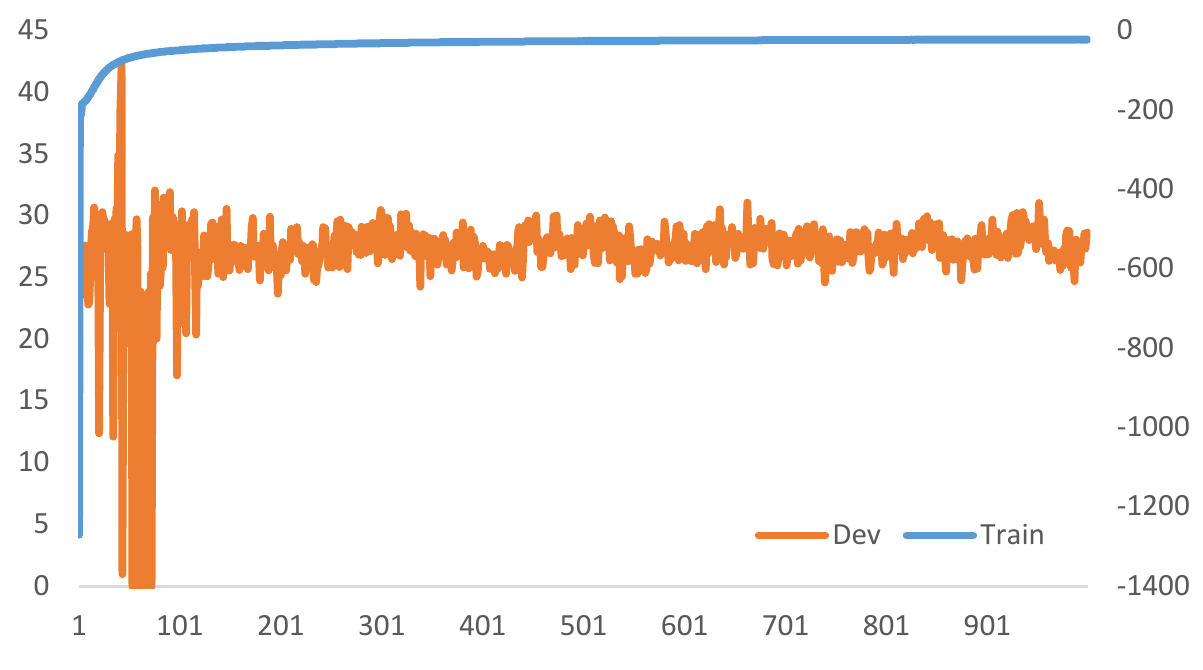}
  	\end{tabular}
  	\caption{\textsc{Com} vs Other}
  \end{subtable} \quad
  \begin{subtable}{.33\textwidth}
  	\centering
  	\begin{tabular}{c}
  		\includegraphics[width=\textwidth]{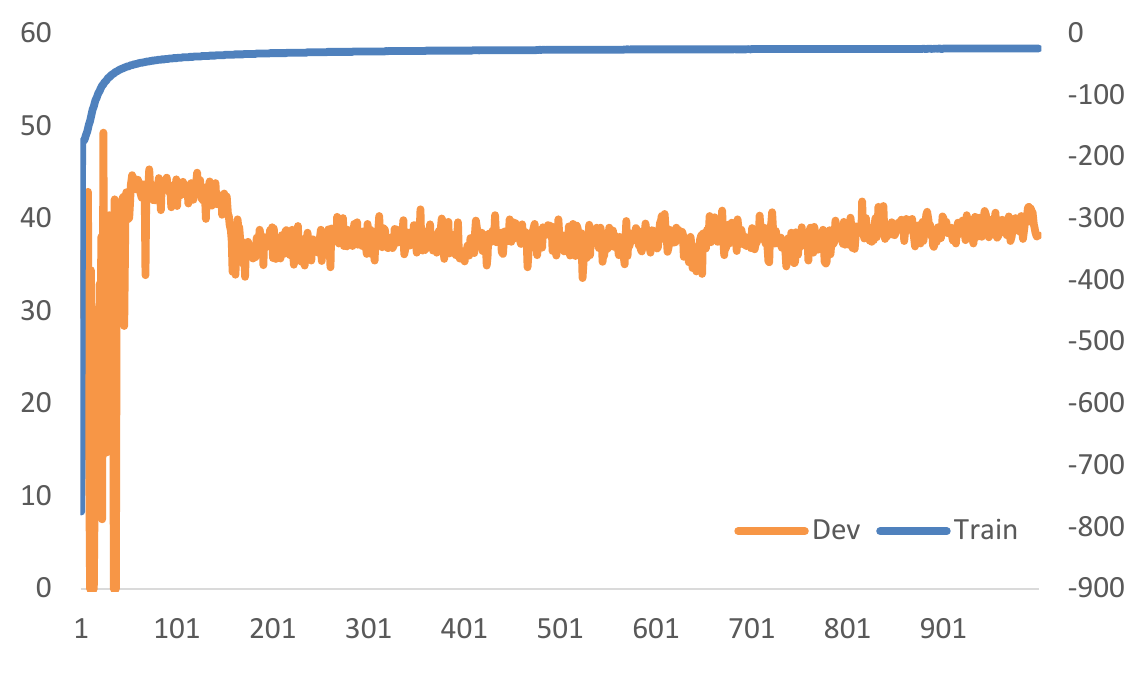}
  	\end{tabular}
  	\caption{\textsc{Con} vs Other}
  \end{subtable}
  \begin{subtable}{.33\textwidth}
  	\centering
  	\begin{tabular}{c}
  		\includegraphics[width=\textwidth]{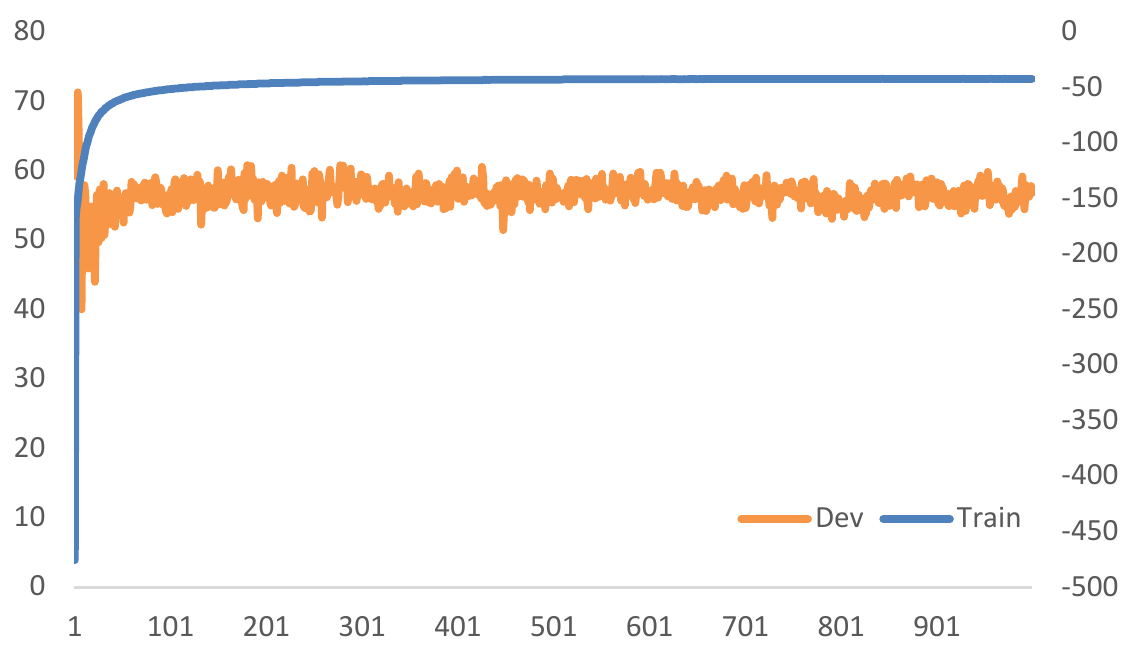}
  	\end{tabular}
  	\caption{\textsc{Exp} vs Other}
  \end{subtable} \quad
  \begin{subtable}{.33\textwidth}
  	\centering
  	\begin{tabular}{c}
  		\includegraphics[width=\textwidth]{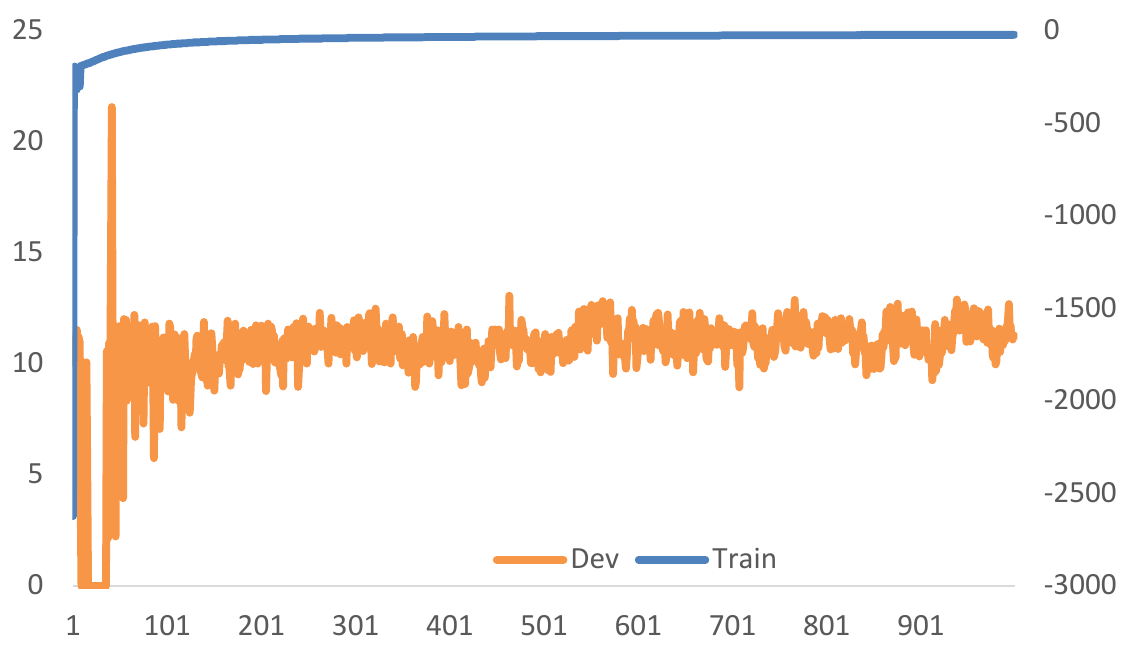}
  	\end{tabular}
  	\caption{\textsc{Tem} vs Other}
  \end{subtable}
  \caption{\label{train_dev} Illustration of the variational lower bound (blue color) on the training set and F-score (brown color) on the development set. Horizontal axis: the epoch numbers; Vertical axis: the F1 score for relation classification (left) and the estimated average variational lower bound per datapoint (right).}
\end{figure*}
 
\section{Experiments}

We conducted experiments on English implicit DRR task to validate the effectiveness of VarNDRR.\footnote{Source code is available at https://github.com/DeepLearnXMU/VarNDRR.} 

\subsection{Dataset}

We used the largest hand-annotated discourse corpus {\it PDTB 2.0}\footnote{http://www.seas.upenn.edu/ pdtb/} \cite{prasad2008penn} (PDTB hereafter). This corpus contains discourse annotations over 2,312 Wall Street Journal articles, and is organized in different sections. Following previous work \cite{pitler-louis-nenkova:2009:ACLIJCNLP,zhou2010predicting,lan2013leveraging,biaozhang:2015:emnlp:drr}, we used sections 2-20 as our training set, sections 21-22 as the test set. Sections 0-1 were used as the development set for hyperparameter optimization.


In PDTB, discourse relations are annotated in a predicate-argument view. Each discourse connective is treated as a predicate that takes two text spans as its arguments. The discourse relation tags in PDTB are arranged in a three-level hierarchy, where the top level consists of four major semantic \emph{classes}: \textsc{Temporal} (\textsc{Tem}), \textsc{Contingency} (\textsc{Con}), \textsc{Expansion} (\textsc{Exp}) and \textsc{Comparison} (\textsc{Com}). Because the top-level relations are general enough to be annotated with a high inter-annotator agreement and are common to most theories of discourse, in our experiments we only use this level of annotations.

We formulated the task as four separate one-against-all binary classification problems: each top level class vs. the other three discourse relation classes. We also balanced the training set by resampling training instances in each class until the number of positive and negative instances are equal. In contrast, all instances in the test and development set are kept in nature. The statistics of various data sets is listed in Table \ref{pdtb_data}.

\subsection{Setup}

We tokenized all datasets using {\it Stanford NLP Toolkit}\footnote{http://nlp.stanford.edu/software/corenlp.shtml}. For optimization, we employed the Adam algorithm~\cite{DBLP:journals/corr/KingmaB14} to update parameters. With respect to the hyperparameters $M,L,A$ and the dimensionality of all vector representations, we set them according to previous work \cite{kingma2014autoencoding,DBLP:conf/icml/RezendeMW14} and preliminary experiments on the development set. Finally, we set $M=16,A=1000,L=1,d_z=20,d_{x_1}=d_{x_2}=10001,d_{h_1}=d_{h_2}=d_{h_1^\prime}=d_{h_2^\prime}=d_m=d_{h_y}=400,d_y=2$ for all experiments.\footnote{There is one dimension in $d_{x_1}$ and $d_{x_2}$ for unknown words.}. All parameters of VarNDRR are initialized by a Gaussian distribution ($\mu=0, \sigma=0.01$). For Adam, we set $\beta_1=0.9$, $\beta_2=0.999$ with a learning rate $0.001$. Additionally, we tied the following parameters in practice: $W_{h_1}$ and $W_{h_2}$, $W_{x_1^{\prime}}$ and $W_{x_2^{\prime}}$.

We compared VarNDRR against the following two different baseline methods:
\begin{itemize}
\item
{\bf SVM:} a support vector machine (SVM) classifier\footnote{http://svmlight.joachims.org/} trained with several manual features. 
\item
{\bf SCNN:} a shallow convolutional neural network proposed by Zhang et al.~\shortcite{biaozhang:2015:emnlp:drr}.
\end{itemize}
We also provide results from two state-of-the-art systems:
\begin{itemize}
\item
{\bf Rutherford and Xue~\shortcite{rutherford-xue:2015:NAACL-HLT}} convert explicit discourse relations into implicit instances.
\item
{\bf Ji and Eisenstein~\shortcite{TACL536}} augment discourse representations via entity connections.
\end{itemize}
Features used in {\bf SVM} are taken from the state-of-the-art implicit discourse relation recognition model, including \emph{Bag of Words}, \emph{Cross-Argument Word Pairs}, \emph{Polarity}, \emph{First-Last, First3}, \emph{Production Rules}, \emph{Dependency Rules} and \emph{Brown cluster pair} \cite{rutherford-xue:2014:EACL}. In order to collect bag of words, production rules, dependency rules, and cross-argument word pairs, we used a frequency cutoff of 5 to remove rare features, following Lin et al.~\shortcite{lin2009recognizing}.

\subsection{Classification Results}

Because the development and test sets are imbalanced in terms of the ratio of positive and negative instances, we chose the widely-used F1 score as our major evaluation metric. In addition, we also provide the precision, recall and accuracy for further analysis. Table \ref{class_result} summarizes the classification results.

From Table \ref{class_result}, we observe that the proposed VarNDRR outperforms {\bf SVM} on \textsc{Com}/\textsc{Exp}/\textsc{Tem} and {\bf SCNN} on \textsc{Exp}/\textsc{Com} according to their F1 scores. Although it fails on \textsc{Con}, VarNDRR achieves the best result on \textsc{EXP}/\textsc{Com} among these three models. Overall, VarNDRR is competitive in comparison with these two baselines. With respect to the accuracy, our model does not yield substantial improvements over the two baselines. This may be because that we used the F1 score rather than the accuracy, as our selection criterion on the development set. With respect to the precision and recall, our model tends to produce relatively lower precisions but higher recalls. This suggests that the improvements of VarNDRR in terms of F1 scores mostly benefits from the recall values. 

Comparing with the state-of-the-art results of previous work \cite{TACL536,rutherford-xue:2015:NAACL-HLT}, VarNDRR achieves comparable results in term of the F1 scores. Specifically, VarNDRR outperforms Rutherford and Xue~\shortcite{rutherford-xue:2015:NAACL-HLT} on \textsc{EXP}, and Ji and Eisenstein~\shortcite{TACL536} on \textsc{TEM}. However, the accuracy of our model fails to surpass these models. We argue that this is because both baselines use many manual features designed with prior human knowledge, but our model is purely neural-based.
 

Additionally, we find that the performance of our model is proportional to the number of training instances. This suggests that collecting more training instances (in spite of the noises) may be beneficial to our model.

\subsection{Variational Lower Bound Analysis}

In addition to the classification performance, the efficiency in learning and inference is another concern for variational methods. Figure \ref{train_dev} shows the training procedure for four tasks in terms of the variational lower bound on the training set. We also provide F1 scores on the development set to investigate the relations between the variational lower bound and recognition performance.

We find that our model converges toward the variational lower bound considerably fast in all experiments (within 100 epochs), which resonates with the previous findings \cite{kingma2014autoencoding,DBLP:conf/icml/RezendeMW14}. However, the change trend of the F1 score does not follow that of the lower bound which takes more time to converge. Particularly to the four discourse relations, we further observe that the change paths of the F1 score are completely different. This may suggest that the four discourse relations have different properties and distributions. 

In particular, the number of epochs when the best F1 score reaches is also different for the four discourse relations. This indicates that dividing the implicit DRR into four different tasks according to the type of discourse relations is reasonable and better than performing DRR on the mixtures of the four relations.

\section{Related Work}

There are two lines of research related to our work: {\it implicit discourse relation recognition} and {\it variational neural model}, which we describe in succession.

{\it Implicit Discourse Relation Recognition} Due to the release of Penn Discourse Treebank \cite{prasad2008penn} corpus, constantly increasing efforts are made for implicit DRR. Upon this corpus, Pilter et al.~\shortcite{pitler-louis-nenkova:2009:ACLIJCNLP} exploit several linguistically informed features, such as polarity tags, modality and lexical features. Lin et al.~\shortcite{lin2009recognizing} further incorporate context words, word pairs as well as discourse parse information into their classifier. Following this direction, several more powerful features have been exploited: entities \cite{louis2010using}, word embeddings \cite{conf/emnlp/BraudD15}, Brown cluster pairs and co-reference patterns \cite{rutherford-xue:2014:EACL}. With these features, Park and Cardie~\shortcite{park2012implicit} perform feature set optimization for better feature combination. 

Different from feature engineering, predicting discourse connectives can indirectly help the relation classification \cite{zhou2010predicting,patterson2013predicting}. In addition, selecting explicit discourse instances that are similar to the implicit ones can enrich the training corpus for implicit DRR and gains improvement \cite{wang2012implicit,lan2013leveraging,braud-denis:2014:Coling,fisher-simmons:2015:ACL-IJCNLP,rutherford-xue:2015:NAACL-HLT}. Very recently, neural network models have been also used for implicit DRR due to its capability for representation learning \cite{TACL536,biaozhang:2015:emnlp:drr}.

Despite their successes, most of them focus on the discriminative models, leaving the field of generative models for implicit DRR a relatively uninvestigated area. In this respect, the most related work to ours is the latent variable recurrent neural network recently proposed by Ji et al.~\shortcite{ji-haffari-eisenstein:2016:N16-1}. However, our work differs from theirs significantly, which can be summarized in the following three aspects: 1) they employ the recurrent neural network to represent the discourse arguments, while we use the simple feed-forward neural network; 2) they treat the discourse relations directly as latent variables, rather than the underlying semantic representation of discourses; 3)  their model is optimized in terms of the data likelihood, since the discourse relations are observed during training. However, VarNDRR is optimized under the variational theory.

{\it Variational Neural Model} In the presence of continuous latent variables with intractable posterior distributions, efficient inference and learning in directed probabilistic models is required. Kingma and Welling~\shortcite{kingma2014autoencoding} as well as Rezende et al.~\shortcite{DBLP:conf/icml/RezendeMW14} introduce variational neural networks that employ an approximate inference model for intractable posterior and reparameterized variational lower bound for stochastic gradient optimization. Kingma et al.~\shortcite{DBLP:conf/nips/KingmaMRW14} revisit the approach to semi-supervised learning with generative models and further develop new models that allow effective generalization from a small labeled dataset to a large unlabeled dataset. Chung et al.~\shortcite{DBLP:journals/corr/ChungKDGCB15} incorporate latent variables into the hidden state of a recurrent neural network, while Gregor et al.~\shortcite{DBLP:journals/corr/GregorDGW15} combine a novel spatial attention mechanism that mimics the foveation of human eyes, with a sequential variational auto-encoding framework that allows the iterative construction of complex images.

We follow the spirit of these variational models, but focus on the adaptation and utilization of them onto implicit DRR, which, to the best of our knowledge, is the first attempt in this respect.

\section{Conclusion and Future Work}

In this paper, we have presented a variational neural discourse relation recognizer for implicit DRR. Different from conventional discriminative models that directly calculate the conditional probability of the relation $\mathbf{y}$ given discourse arguments $\mathbf{x}$, our model assumes that it is a latent variable from an underlying semantic space that generates both $\mathbf{x}$ and $\mathbf{y}$. In order to make the inference and learning efficient, we introduce a neural discourse recognizer and two neural latent approximators as our generative and inference model respectively. Using the reparameterization technique, we are able to optimize the whole model via standard stochastic gradient ascent algorithm. Experiment results in terms of classification and variational lower bound verify the effectiveness of our model.

In the future, we would like to exploit the utilization of discourse instances with explicit relations for implicit DRR. For this we can start from two directions: 1) converting explicit instances into pseudo implicit instances and retraining our model; 2) developing a semi-supervised model to leverage semantic information inside discourse arguments. Furthermore, we are also interested in adapting our model to other similar tasks, such as nature language inference.

\section*{Acknowledgments}
The authors were supported by National Natural Science Foundation of China (Grant Nos 61303082, 61672440, 61402388, 61622209 and 61403269), Natural Science Foundation of Fujian Province (Grant No. 2016J05161), Natural Science Foundation of Jiangsu Province (Grant No. BK20140355), Research fund of the Provincial Key Laboratory for Computer Information Processing Technology in Soochow University (Grant No. KJS1520), and Research fund of the Key Laboratory for Intelligence Information Processing in the Institute of Computing Technology of the Chinese Academy of Sciences (Grant No. IIP2015-4). We also thank the anonymous reviewers for their insightful comments.

\bibliography{emnlp2016}
\bibliographystyle{emnlp2016}

\end{document}